\pdfoutput=1
\documentclass[11pt]{article}

\usepackage[]{acl}

\usepackage{times}
\usepackage{latexsym}

\usepackage[T1]{fontenc}
\usepackage[utf8]{inputenc}
\usepackage{hyperref}       % hyperlinks
\usepackage{url}            % simple URL typesetting
\usepackage{booktabs}       % professional-quality tables
\usepackage{amsfonts}       % blackboard math symbols
\usepackage{nicefrac}       % compact symbols for 1/2, etc.
\usepackage{microtype}      % microtypography
\usepackage{xcolor}         % colors
\usepackage{subfigure}
\usepackage{graphicx}
\usepackage{amsmath}
\usepackage{amsfonts} 
\usepackage{multirow}
\usepackage{geometry}
\usepackage{mathrsfs}
\usepackage{enumerate}
\usepackage{longtable}
\usepackage{amsthm}
\usepackage{wrapfig}
\usepackage{colortbl}
\usepackage{caption}
\usepackage{tabularx} % for 'tabularx' env. and 'X' col. type
\usepackage{ragged2e} % for \RaggedRight macro
\usepackage{booktabs} % for \toprule, \midrule etc macros
\newcolumntype{L}{>{\RaggedRight\hangafter=1\hangindent=0em}X}
\usepackage{microtype}
\definecolor{mydarkgreen}{RGB}{0,128,0}

\title{Chain of History: Learning and Forecasting with LLMs for Temporal Knowledge Graph Completion}

\author{
  Ruilin Luo$^{1}$\thanks{~{Equal contribution.~$\dagger$~Corresponding author: yang.yujiu@sz.tsinghua.edu.cn.}},~
  Tianle Gu$^{1\ast}$,~
  Haoling Li$^{1\ast}$,~
  Junzhe Li$^{2}$,~
  Zicheng Lin$^{1}$,~
  Jiayi Li$^{3}$,~
  Yujiu Yang$^{1\dagger}$ \\
  \\$^{1}$Tsinghua Shenzhen International Graduate School, Tsinghua University
  \\$^{2}$School of Computer Science, Peking University
  \\$^{3}$Baidu Inc.
  \\ \texttt{\{lrl23,gtl23,li-hl23\}@mails.tsinghua.edu.cn}
}

\begin{document}
\maketitle
\begin{abstract}
Temporal Knowledge Graph Completion (TKGC) is a complex task involving the prediction of missing event links at future timestamps by leveraging established temporal structural knowledge. This paper aims to provide a comprehensive perspective on harnessing the advantages of Large Language Models (LLMs) for reasoning in temporal knowledge graphs, presenting an easily transferable pipeline. In terms of graph modality, we underscore the LLMs' prowess in discerning the structural information of pivotal nodes within the historical chain. As for the generation mode of the LLMs utilized for inference, we conduct an exhaustive exploration into the variances induced by a range of inherent factors in LLMs, with particular attention to the challenges in comprehending reverse logic. We adopt a parameter-efficient fine-tuning strategy to harmonize the LLMs with the task requirements, facilitating the learning of the key knowledge highlighted earlier. Comprehensive experiments are undertaken on several widely recognized datasets, revealing that our framework exceeds or parallels existing methods across numerous popular metrics. Additionally, we execute a substantial range of ablation experiments and draw comparisons with several advanced commercial LLMs, to investigate the crucial factors influencing LLMs' performance in structured temporal knowledge inference tasks.
\end{abstract}

\section{Introduction}

Knowledge Graphs~(KGs), defined as meticulously structured repositories of deterministic knowledge, have been utilized across a wide range of domains such as recommender systems~\cite{app_recom}, question-answering~\cite{app_qa}, and more recently, in the emerging field of Retrieval-augmented Generation~(RAG)~\cite{app_rag,retrieval}. In recent years, the concept of Temporal Knowledge Graphs~(TKGs) has gained increased attention due to their ability to provide more accurate information.~\cite{ttranse,xerte,tirgn,icl_tkgc}. A Temporal Knowledge Graph~(TKG) stores numerous facts in the form of quadruples $(e_h, r, e_t, t_T)$, denoting that $e_h$ has a directional edge $r$ into $e_t$ at timestamp $t_T$. Given a series of observed facts denoted as $\mathcal{F}=\{(s,p,o, t_s)|s, o \in \mathcal{S}, p\in \mathcal{P}, t_s < T\}$, TKGC under extrapolative setting requires the capability to predict links to future timestamps, i.e., quadruples containing $t_s\geq T$. This extrapolative setting has attracted more research than the interpolation setting, which primarily focuses on events in observed timestamps~\cite{cygnet,titer}.

\begin{figure}[!t]
    \centering
    \includegraphics[width=\linewidth]{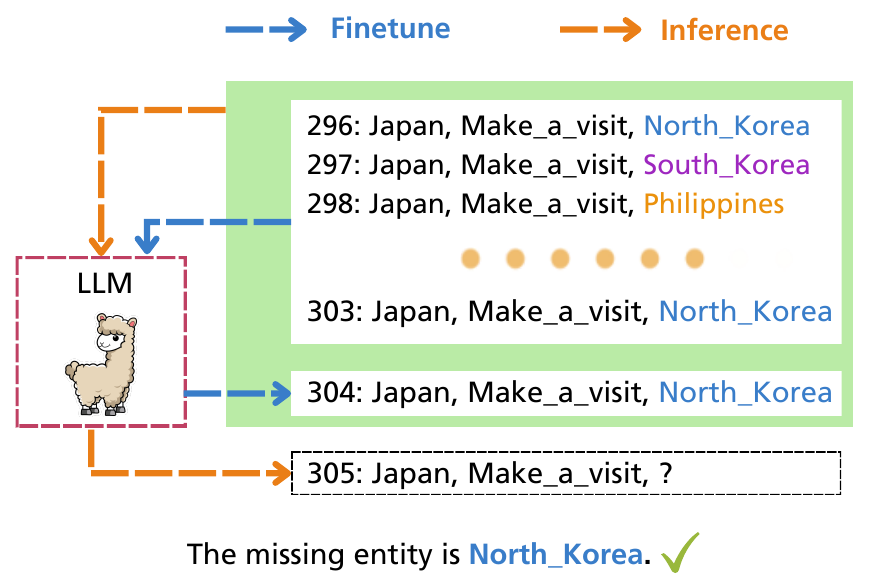}
    \caption{LLM undergoes fine-tuning on known data and subsequently utilizes the chain of known factual information to generate the next event.}
    \label{fig: teaser graph}
    \vspace{-0.3cm}
\end{figure}

Previous research has approached the TKGC task from various angles. Some models, integrating Graph Neural Networks (GNNs) with gated mechanisms, focus on the evolution of embeddings over time~\cite{recurrent,regcn,tirgn,hgls}. Rule learning aims to provide ample prior knowledge~\cite{tlogic}, while reinforcement learning models~\cite{titer} propose time-shaped rewards to guide the learning process. Despite these efforts, these methods often fall short in utilizing the rich text information and underperform when the links are sparse. Recently, with the demonstrated capabilities of LLMs in various fields, some attempts have been made to explore the utilization of LLMs for TKGC tasks. ~\cite{icl_tkgc} explores the potential of in-context learning~(ICL) capabilities of LLMs to perform on the TKGC task. GenTKG~\cite{gentkg} leverages the partial idea of tLogic~\cite{tlogic} to provide LLMs with the most temporal logic-relevant inputs to counsel decisions. 

In this paper, we seek to thoroughly examine whether LLMs are effective TKG reasoning agents and how to reveal genuinely beneficial factors.  On one hand, TKGs are essentially graph structures with textual information, and recent research has demonstrated that LLMs possess certain capabilities in understanding structural information, yielding promising results in tasks such as node classification~\cite{graphgpt,distag,gpt4graph,one4all}. On the other hand, as an inference task, TKGC specifically requires the natural advantage of textual reasoning possessed by LLMs. Considering the aforementioned characteristics, we develop a general and easily transferable framework: 1) For structural awareness of TKGs, in addition to considering the history that directly provides candidate answers, we also incorporate additional neighboring interaction information of entities and relations. 2) Regarding LLM inference within the TKG context, our focus lies in mitigating the reversal curse in structured expression reasoning. 3) We employ the Parameter-Efficient Fine-Tuning~(PEFT) technique for fine-tuning LLMs to enhance the model's understanding of historical context and integrate the two aforementioned solutions.

Specifically, as shown in Fig.~\ref{fig: teaser graph}, during the fine-tuning process, we partition the known data into an input section and a supervised labeling segment, guiding LLMs in adapting the mapping relationship between the textual information of the specific TKG and the intricate logic inherent in temporal events. We propose to use local information across multiple single-step graphs for historical data augmentation to explore the ability of LLMs to perceive graph-modality information. In addition, we explore different ways of reverse data incorporation to alleviate the reversal curse~\cite{lv2023curse} problem in structured knowledge reasoning.

We carry out comprehensive experiments on widely used TKGC datasets, including the ICEWS~\cite{regcn} series from news and the commonsense dataset YAGO~\cite{yago3}. Significantly, we report the Hits@n metric under raw setting and time-aware filtered setting, achieving highly competitive results. We also provide the 8-shot ICL\footnote{ Prompts can be found in Appendix~\ref{sec:app coh} and~\ref{sec:app icl}.} performance of several open-source models as a comparative reference. Furthermore, we conduct exhaustive ablation experiments to validate the effectiveness of structure-based historical data augmentation methods and the introduction of reverse logic. Additionally, we investigate the impact of historical chain length, model size, and the performance of LLMs like GPT-4 and GPT-3.5-turbo, with the aim to uncover key factors influencing temporal structural information reasoning using LLMs.

\section{Related Work}

\textbf{Temporal Knowledge Graph Completion} involves two essential reasoning settings: interpolation and extrapolation. Interpolation-based TKG reasoning addresses the challenge of filling in missing links within observed timestamps. TTransE~\cite{ttranse} introduces time-based encoding through translation operations. TNTComplEx~\cite{tnt_complex} and TuckERTNT~\cite{tnt_tucker} propose complex decomposition and TuckER decomposition of four-order tensors, respectively, to augment model expressiveness under temporal conditions. However, the interpolation setting has limitations, as it cannot infer missing information in future timestamps, thereby restricting its applicability.

Extrapolative reasoning in TKGC, involving the prediction of facts for future timestamps, represents a more challenging yet valuable task. Recent works have concentrated on leveraging multi-relational graph convolutional networks~\cite{regcn, renet}. xERTE~\cite{xerte} captures query-related subgraph information through dynamic pruning operations. TANGO~\cite{tango} adopts neural ordinary differential equations to model the temporal representation of entities. TITer~\cite{titer} stands out as the first model to utilize temporal-path-based reinforcement learning for TKG reasoning. TLogic~\cite{tlogic} enhances interpretability by extracting temporal logic rules through random exploration of time. TiRGN~\cite{tirgn} and HGLS~\cite{hgls} utilize graph learning methods for comprehensive structural information capture during temporal wandering. ~\cite{icl_tkgc} first explores the potential of ICL in TKGC. GenTKG~\cite{gentkg} provides the most relevant interactions in temporal logic for LLMs to learn and infer.
 
\noindent\textbf{LLMs-as-Predictors}
Many recent studies transform graph structure information into sequential representations and utilize LLMs as standalone predictors. Graph4GPT~\cite{guo2023gpt4graph} uses InstructGPT-3~\cite{ouyang2022training} to conduct an empirical study to assess LLMs' capabilities in graph understanding, and GraphLLM~\cite{chai2023graphllm} uses LLaMA2 for the graph reasoning task, but these work ignore LLM's ability to TKGC. Most relevant to our work,~\cite{lee2023temporal} uses ICL with LLMs for TKGC, which may not fully exploit the extensive learning capabilities of LLMs.

\noindent\textbf{Parameter-Efficient Fine-tuning}
Recent studies have introduced several PEFT techniques, including the addition of adapters~\cite{he2022unified,rebuffi2017learning,houlsby2019parameterefficient,bapna2019simple}, which entail the insertion of small trainable feed-forward networks between fixed pre-trained models. Additionally, low-rank updates~\cite{hu2021lora} have been proposed as an alternative, wherein the fine-tuning process leverages low-dimensional representations. Moreover, prompt tuning~\cite{lester2021power} and prefix tuning~\cite{li-liang-2021-prefix} have been developed, which involve augmenting the model's input or activations with learnable parameters.

\section{Preliminary}

\textbf{Definition 3.1.~TKGC} A TKG is defined as a sequence $\mathcal{G}=\{\mathcal{G}_1, \cdots, \mathcal{G}_t, \cdots, \mathcal{G}_n \}$ comprising static KGs. Here, each static KG denoted as $\mathcal{G}_{t}$ contains factual triplets at timestamp $t$. A single static KG is formulated as $\{\mathcal{E}, \mathcal{R}, \mathcal{T}\}$, in which $\mathcal{E}$, $\mathcal{R}$ and $\mathcal{T}=\{s_i,p_j,o_k\}$ respectively represent entities, relations and triplets within it. TKGC involves bidirectional prediction of query quadruples, specifically, $(s_i, p_j, ?, t_s)$ and $(o_k, p_j^{-1}, ?, t_s)$.

\noindent\textbf{Definition 3.2.~Fine-tuning} Given a pre-trained LLM denoted as $\mathcal{M}$ with parameters $\mathcal{\theta}$, and a dataset comprising $n$ instances $\{\text{Query}_{i}, \text{Response}_{i}\}$, the fine-tune processing aims to minimize the following loss function:
\begin{equation}
\boldsymbol{\theta}^{\star}=\arg \min _{\boldsymbol{\theta}^{\prime}} \sum_{i=0}^{n-1} \mathcal{L}\left(\mathcal{M}\left(\mathcal{Q} | \boldsymbol{\theta}^{\prime}\right) ; \mathcal{R}\right)
\end{equation}
where  $\mathcal{M}(|\theta')$ denotes the output of the fine-tuned LLM $\mathcal{M}$ with parameters $\theta'$, $\mathcal{Q}$ represents Query and $\mathcal{R}$ represents response.

\section{Methodology}
\subsection{Structure-augmented History Modeling}

The LLM's predictions of undiscovered links in the TKG rely on knowledge derived from historical facts. In particular, when dealing with a query quadruple represented as $q=(s_i, p_j, ?, t_q)$ in a forward reasoning mode, we aim to model the historical chain $\mathcal{H}_q$ associated with this query.

\noindent\textbf{Schema-matching History.} \hspace*{0.3mm} The initial set of historical facts we leverage originates from schema-matching records, denoted as $H_s=\{(s_i,p_j,o,t) | o\in \mathcal{E}, t < t_q\}$. Specifically, given a query (\textit{Japan, Make\_a\_visit, ?, 305}), $\mathcal{H}_s=\{(\textit{Japan, Make\_a\_visit, North\_Korea, 296}), \cdots, \\ (\textit{Japan, Make\_a\_visit, North\_Korea, 304})\}$ encompasses relevant schema-matching facts that align with the subject and predicate of the query $q$, providing inference basis for LLMs.

\noindent\textbf{Entity-augmented History.} \hspace*{0.3mm} Similar to many prior works that leverage structural information from KGs to enhance the reasoning capabilities of LLMs~\cite{luo2023reasoning, tian2023reasoning}, we focus on semantically enriching the representation of central entities by utilizing links with neighbors in TKGs. The entity-augmented history $\mathcal{H}_{e}$ is defined as $\{(s_i, p, o, t)| (s_i,p,o,t)\in \mathcal{G}_t, p\in \mathcal{R}, o \in \mathcal{E}, t< t_q\}$ formally. 

\noindent\textbf{Relation-augmented History.} \hspace*{0.3mm} In addition to completing the historical chain based on entity-based neighbor information, we introduce a supplementary strategy based on relations. We believe that it's beneficial for enhancing the model's intrinsic understanding of relation inference~\cite{oneshot_relation_learning}. Formally, relation-augmented history set $\mathcal{H}_r=\{(s,p_j, o, t)|(s,p_j,o,t)\in \mathcal{G}_{t}, p,o\in \mathcal{E}, t<t_q\}$.

When modeling $\mathcal{H}_q$, we adhere to two criteria for selecting data from $\mathcal{H}_s,\mathcal{H}_e$, and $\mathcal{H}_r$. \romannumeral1) We prioritize the ground-truth history directly related to $q$, which is $\mathcal{H}_s$. If the history length does not meet the specified value, we then sequentially incorporate facts from $\mathcal{H}_e$ and $\mathcal{H}_r$. \romannumeral2) Data close to the current timestamp is introduced with priority. By following these two criteria, we aim to select the most relevant knowledge to inspire forecasting capabilities in LLMs.

\subsection{Introduction of Reverse Logic}
Similar to reasoning on static KGs, we require the model to also possess the capability of reverse inference on TKG~\cite{regcn}. However, recent research indicates that LLM's reasoning has encountered the issue of reversal curse~\cite{qi2023curse,lucas2023curse,lv2023curse}. In this problem, models often succeed in correctly deducing questions like '\textit{Who is Tom Cruise's mother?}' but struggle to answer '\textit{Who is the son of Mary Lee Pfeiffer?}'. We believe that this phenomenon also exists in structured knowledge reasoning. We propose using three prompt strategies to incorporate reverse quadruples during the fine-tuning phase to alleviate this issue, and explore the performance patterns in the context of structured knowledge reasoning scenarios. 

As demonstrated in Tbl.~\ref{tab: reverse_prompts}, the most ordinary construction is to treat the structure of backward inferences as forward inferences. The text-aware prompt leverages \textit{reverse} to indicate reverse reasoning, and the position-aware prompt follows the order of backward inference, providing different head entities in the historical records.

\begin{table}[t]
    \centering
    \resizebox{0.95\columnwidth}{!}{
        \begin{tabular}{c|l}
            \hline
            Strategy & Prompt \\
            \hline
            \multirow{5}{*}{Ordinary} & 280: [Japan, Make\_a\_visit, China] \\
             & 281: [Japan, Make\_a\_visit, Vietnam] \\
             & $\cdots$ \\
             & 304: [Japan, Make\_a\_visit, Kiichi\_Miyazawa] \\
             & Query: 305: [Japan, Make\_a\_visit, ] \\
             \hline
            \multirow{5}{*}{Text-aware} & 280: [Japan, reverse Make\_a\_visit, China] \\
             & 281: [Japan, reverse Make\_a\_visit, Vietnam] \\
             & $\cdots$ \\
             & 304: [Japan, reverse Make\_a\_visit, Kiichi\_Miyazawa] \\
             & Query: 305: [Japan, reverse Make\_a\_visit, ] \\
            \hline
            \multirow{5}{*}{Position-aware} & 280: [China, Make\_a\_visit, Japan] \\
             & 281: [Vietnam, Make\_a\_visit, Japan] \\
             & $\cdots $\\
             & 304: [Kiichi\_Miyazawa, Make\_a\_visit, Japan] \\
             & Query: 305: [ , Make\_a\_visit, Japan] \\
             \hline
        \end{tabular}
    }
    \vspace{-0.1cm}
    \caption{Prompts for query (\textit{Japan, Make\_a\_visit$^{-1}$, ?, 305}).}
    \label{tab: reverse_prompts}
    \vspace{-0.3cm}
\end{table}

\subsection{Instruction-tuning in TKGC}
Instruction-tuning~\cite{wei2021finetuned} achieves remarkable zero-shot generalization results by training LLMs on different tasks with instructions. While prior work has demonstrated the effectiveness of fine-tuning LLMs via full-parameter updates, this approach presents considerable challenges at large scale. Hence, we apply the Low-Rank Adaptation~(LoRA)~\cite{hu2021lora} method due to its effectiveness for Llama-style models. This method, founded on the plugin encapsulation strategy of PEFT, furnishes us with lightweight task-specific plugins. %Speficially, given a large language model $\mathcal{M}$, the instruction $\mathcal{Q}$ and the answer of LLM $\hat{\mathcal{R}=M(Q)}$ are formulated as%

The LLM $\mathcal{M}$ generates a sequence of tokens $\hat{\mathcal{R}}=\{\hat{r_1}, \hat{r_2},...\hat{r_n}\}$, where response $\mathcal{R}$ we need must be extracted and consists of a set of consecutive tokens. Similarly to most fine-tuning LLMs process using LoRA, the parameter update for a pre-trained weight matrix ${W_{0}} \in \mathbb{R}^{d \times k}$ is specified by product of two low-rank matrices $W_A$ and $W_B$:
\begin{equation}
    \delta W=W_AW_B
\end{equation}
where ${W_{A}} \in \mathbb{R}^{d \times r}$ and ${W_{B}} \in \mathbb{R}^{r \times k}$ are matrices of trainable parameters and $\operatorname{rank} r \ll \min (d, k)$. Therefore, the forward pass for $h=W_0x$ is altered as :
\begin{equation}
    h = W_0{x}+\delta W_x=W_0 x+W_A W_B x
\end{equation}
We employ cross-entropy loss which constrains the similarity between estimated and ground-truth tokens, to fine-tune LLMs by LoRA, which can be presented as
\begin{equation}
    \mathcal{L} = CE(\mathcal{\hat{R}}, \tilde{R})
\end{equation}
where $\hat{\mathcal{R}}$ is the temporal knowledge graph completion predicted by LLM $\mathcal{M}$ and $\tilde{R}$ is the given label.

\subsection{Predict with LLMs}
The instructions constructed are fed into the trained LLMs for prediction. The response is obtained by beam search, which is a decoding strategy that maintains $k$ beams of possible generated responses at each time step $t$. The generation of response is updated as follows: for each generated response, the $k$ tokens with the highest probabilities are selected based on Eq.~\ref{maxtoken}. This results in $k \times k$ new response candidates. The next $k$ beams of response are obtained by selecting the top $k$ responses with the highest probabilities from the generated response candidates. The highest probability is determined by the product of probabilities of $|\mathcal{\hat{R}}|$ tokens that constitute the response, where $|\mathcal{\hat{R}}|$ represents the length of the current response.

\begin{equation}
    r_t = argmax_{r}P(r|r_{1:t-1})
    \label{maxtoken}
\end{equation}

In this context, the single step setting is employed, wherein for each test query in the test dataset, the model can access the ground truth from past timestamps. Consequently, after the prediction for this step is completed, the ground truth from the current timestamp is added to the history of the next timestamp before its execution.

\section{Experiments}
\subsection{Datasets}

In our experimental setup, we utilize the ICEWS14 dataset~\cite{ta_distmult}, ICEWS18 dataset~\cite{regcn}, ICEWS05-15 dataset~\cite{icews0515}, and YAGO dataset~\cite{yago3} as benchmarks for evaluation. The specific statistics are listed in Tbl.~\ref{tab: statistics}. We employ partition criteria widely accepted in prior studies~\cite{xerte} and establish instruction-tuning data on the validation set. Specifically, for the ordered timestamp set $T=\{t_{train}^1, t_{train}^2,\cdots, t_{train}^n, t_{val}^1, \cdots, t_{val}^m\}$, comprising training and validation sets, when gathering historical data for timestamp $t_{val}^i$, we observe only facts within the range $t<t_{val}^i$. In the context of testing under a single-step setup~\cite{know-evolve}, for a query at timestamp $t_q$, we construct a ground-truth chain of history based on facts preceding timestamp $t_q$, serving as the input to the model. 
\begin{table}[htbp]
\vspace{-0.3cm}
    \begin{center}
      \resizebox{1.0\columnwidth}{!}{
    \begin{tabular}{lcccccc} 
      \hline % <-- Toprule here
      \textbf{Datasets} & \textbf{Entity} & \textbf{Relation} & \textbf{Train} & \textbf{Valid} & \textbf{Test} & \textbf{Interval} \\
      \hline
      ICEWS14 & 6869 & 230 & 74845 & 8514 & 7371 & 1\ day \\
      ICEWS05-15 & 10094 & 251 & 368868 & 46302 & 46159 & 1\ day \\
      ICEWS18 & 23033 & 256 & 373018 & 45995 & 49545 & 1\ day \\
      YAGO & 10623 & 10 & 161540 & 19523 & 20026 & 1\ year \\
      \hline
    \end{tabular}}
    \caption{Statistics of leveraged datasets. 
    }
    \label{tab: statistics}
    \end{center}
    \vspace{-0.3cm}
\end{table}

\begin{table*}[!t]
    \begin{center}
      \resizebox{1.0\textwidth}{!}{
    \begin{tabular}{l|ccc|ccc|ccc|ccc} 
      \hline % <-- Toprule here
      \multicolumn{1}{l}{Datasets} & \multicolumn{3}{|c}{\textbf{YAGO}} & \multicolumn{3}{|c}{\textbf{ICEWS14}} &  \multicolumn{3}{|c}{\textbf{ICEWS05-15}} & \multicolumn{3}{|c}{\textbf{ICEWS18}} \\
      \hline 
       Model & Hits@1 & Hits@3 & Hits@10 & Hits@1 & Hits@3 & Hits@10 & Hits@1 & Hits@3 & Hits@10 & Hits@1 & Hits@3 & Hits@10  \\
       \hline 
       % RE-GCN &  0.313	&0.473	&0.626	&0.223	&0.367	&0.525 \\ 
            % &RE-GCN &  0.313	&0.473	&0.626	&0.223	&0.367	&0.525 &0.084	&0.171	&0.299 &0.468	&0.607	&0.729 \\
            RE-NET~\cite{renet} & 0.404 & 0.530 & 0.629 & 0.293 & 0.431 & 0.575 & 0.334 & 0.478 & 0.611 & 0.192 & 0.323 & 0.483 \\
            RE-GCN~\cite{regcn} & 0.499 & 0.663 & 0.779 & 0.297 & 0.441 & 0.586 & 0.336 & 0.487 & 0.658 & 0.193 & 0.331 & 0.494 \\
            xERTE~\cite{xerte} & 0.506 & 0.719 & 0.828 & 0.312 & 0.453 & 0.570 & 0.347 & 0.497 & 0.633 & 0.206 & 0.330 & 0.458 \\
            TANGO$\dagger$~\cite{tango} & 0.409 & 0.554 & 0.637 & 0.151 & 0.272 & 0.431 & 0.311 & 0.476 & 0.622 & 0.178 & 0.314 & 0.460 \\
            Timetraveler~\cite{titer} & 0.494 & 0.675 & 0.790 & 0.313 & 0.451 & 0.571 & 0.341 & 0.494 & 0.667 & 0.210 & 0.325 & 0.437 \\ \hline
            TLogic~\cite{tango} & 0.454 & 0.703 & 0.782 & 0.322 & \text{0.470} & 0.603 & 0.345 & 0.525 & 0.673 & 0.205 & 0.339 & 0.484 \\
      \hline 
            TiRGN~\cite{tirgn} & 0.509 & 0.710 & 0.864 & 0.313 & 0.468 & 0.612 & 0.358 & \textbf{0.535} & 0.690 & 0.202 & 0.350 & 0.514 \\
            HGLS~\cite{hgls} & 0.508 & 0.721 & 0.866 & \textbf{0.349} & \textbf{0.480} & \textbf{0.688} & 0.351 & 0.521 & 0.673 & 0.192 & 0.323 & 0.494 \\ 
      \hline 
      GenTKG~\cite{gentkg} & 0.520 & 0.731 & \underline{0.870} & \textbf{0.349} & \underline{0.473} & 0.619 & 0.360 & 0.525 & 0.687 & \underline{0.215} & \textbf{0.366} & 0.496 \\
      GPT-NeoX-20B-ICL~\cite{neox} & 0.520 & 0.722 & \underline{0.870} & 0.295 & 0.406 & 0.475 & 0.348 & 0.497 & 0.586 & 0.177 & 0.290 & 0.385 \\
       Llama-2-7b-ICL~\cite{llama} & 0.517 & 0.725 & 0.868 & 0.275 & 0.391 & 0.453 & 0.353 & 0.490 & 0.563 & 0.177 & 0.295 & 0.364 \\
       Vicuna-7b-ICL~\cite{vicuna} & 0.514 & 0.714 & 0.868 & 0.270 & 0.386 & 0.453 & 0.347 & 0.483 & 0.563 & 0.172 & 0.288 & 0.364 \\
      \hline 
       Llama-2-7b-CoH & \underline{0.527} & \underline{0.747} & \textbf{0.874} & \underline{0.338} & 0.462 & 0.587 & \underline{0.370} & \underline{0.531} & \underline{0.699} & \textbf{0.219} & \underline{0.361} & \underline{0.520} 
       \\
       Vicuna-7b-CoH & \textbf{0.530} & \textbf{0.754} & 0.859 & 0.315 & 0.445 & \underline{0.648} & \textbf{0.372} & \underline{0.531} & \textbf{0.701} & 0.206 & 0.344 & \textbf{0.531} \\
      \hline 
    \end{tabular}}
    \caption{Temporal forecasting with \textbf{raw} metrics Hits@1, Hits@3 and Hits@10. The best results are highlighted in \textbf{bold} and the second-rank results are \underline{underlined}. The results of the model with $\dagger$ are derived from~\cite{tango}, while other models have been reproduced by us.  %The standard errors of the fusion models are also provided.
    }
    \label{tab: link prediction results PART1}
    \end{center}
\end{table*}

\begin{table*}[!t]
    \begin{center}
      \resizebox{1.0\textwidth}{!}{
    \begin{tabular}{l|ccc|ccc|ccc|ccc} 
      \hline % <-- Toprule here
      \multicolumn{1}{l}{Datasets} & \multicolumn{3}{|c}{\textbf{YAGO}} & \multicolumn{3}{|c}{\textbf{ICEWS14}} &  \multicolumn{3}{|c}{\textbf{ICEWS05-15}} & \multicolumn{3}{|c}{\textbf{ICEWS18}}\\
      \hline 
       Model & Hits@1 & Hits@3 & Hits@10 & Hits@1 & Hits@3 & Hits@10 & Hits@1 & Hits@3 & Hits@10 & Hits@1 & Hits@3 & Hits@10  \\
       \hline 
       % RE-GCN &  0.313	&0.473	&0.626	&0.223	&0.367	&0.525 \\ 
            % &RE-GCN &  0.313	&0.473	&0.626	&0.223	&0.367	&0.525 &0.084	&0.171	&0.299 &0.468	&0.607	&0.729 \\
            RE-NET$\dagger$~\cite{renet} & 0.586 & 0.715 & 0.868 & 0.301 & 0.440 & 0.582 & 0.336 & 0.488 & 0.627 & 0.197 & 0.326 & 0.485 \\
            RE-GCN$\dagger$~\cite{regcn} & 0.788 & 0.843 & 0.886 & 0.313 & 0.470 & 0.613 & 0.366 & 0.527 & 0.671 & 0.215 & 0.354 & 0.515 \\
            xERTE$\dagger$~\cite{xerte} & 0.801 & 0.880 & 0.898 & 0.327 & 0.457 & 0.573 & 0.378 & 0.523 & 0.639 & 0.210 & 0.335 & 0.465 \\
            TANGO$\ddagger$~\cite{tango} & 0.590 & 0.646 & 0.677 & 0.272 & 0.408 & 0.550 & 0.344 & 0.499 & 0.640 & 0.191 & 0.318 & 0.462 \\
            Timetraveler$\dagger$~\cite{titer} & 0.801 & 0.900 & 0.903 & 0.327 & 0.465 & 0.584 & 0.383 & 0.527 & 0.649 & \underline{0.221} & 0.335 & 0.448\\ \hline
            TLogic$\ddagger$~\cite{tango} & 0.740 & 0.789 & 0.791 & 0.336 & 0.483 & 0.612 & 0.362 & 0.531 & 0.674 & 0.205 & 0.340 & 0.485\\
      \hline 
            TiRGN~\cite{tirgn} & 0.839 & 0.907 & 0.923 & 0.328 & 0.481 & 0.622 & 0.379 & \underline{0.544} & 0.698 & 0.220 & \underline{0.366} & \underline{0.522} \\
            HGLS~\cite{hgls} & 0.827 & \underline{0.911} & \underline{0.926} & \textbf{0.368} & \textbf{0.490} & \textbf{0.691} & 0.360 & 0.525 & 0.678 & 0.200 & 0.316 & 0.494 \\ 
      \hline 
      GenTKG~\cite{gentkg} & 0.813 & 0.901 & 0.922 & \underline{0.365} & \underline{0.488} & 0.633 & 0.378 & 0.541 & 0.692 & 0.220 & \textbf{0.370} & 0.497 \\
      GPT-NeoX-20B-ICL~\cite{neox} & 0.792 & 0.890 & 0.909 & 0.295 & 0.406 & 0.475 & 0.367 & 0.503 & 0.587 & 0.192 & 0.300 & 0.389 \\
       Llama-2-7b-ICL~\cite{llama} & 0.767 & 0.852 & 0.868 & 0.286 & 0.397 & 0.453 & 0.353 & 0.490 & 0.563 & 0.177 & 0.294 & 0.364 \\
       Vicuna-7b-ICL~\cite{vicuna} & 0.747 & 0.840 & 0.868 & 0.281 & 0.391 & 0.453 & 0.347 & 0.483 & 0.563 & 0.172 & 0.288 & 0.364 \\
      \hline 
       Llama-2-7b-CoH & \textbf{0.880} & \textbf{0.929} & \textbf{0.931} & 0.349 & 0.470 & 0.591 & \underline{0.386} & 0.541 & \underline{0.699} & \textbf{0.223} & 0.363 & \underline{0.522} 
       \\
       Vicuna-7b-CoH & \underline{0.851} & 0.903 & 0.918 & 0.328 & 0.457 & \underline{0.656} & \textbf{0.392} & \textbf{0.546} & \textbf{0.707} & 0.209 & 0.347 & \textbf{0.536} \\
      \hline 
    \end{tabular}}
    \caption{Temporal forecasting with \textbf{time-aware filtered} metrics Hits@1, Hits@3 and Hits@10. The best results are highlighted in \textbf{bold} and the second-rank results are \underline{underlined}. The results of the model with $\dagger$ are derived from~\cite{tirgn}, and results with $\ddagger$ are taken from~\cite{icl_tkgc}.  %The standard errors of the fusion models are also provided.
    }
    \label{tab: link prediction results PART2}
    \end{center}
\end{table*}

\subsection{Baseline Models}

The models selected for comparative analysis primarily fall into two categories: embedding-based methods and LLM-based approaches. Within the realm of embedding-based methods, we present the performance evaluations of RE-NET~\cite{renet}, RE-GCN~\cite{regcn}, TiRGN~\cite{tirgn}, xERTE~\cite{xerte}, TANGO~\cite{tango}, Timetraveler~\cite{titer}. As for GNN-based methodologies, we choose TiRGN~\cite{tirgn} and HGLS~\cite{hgls} for comparison. Regarding LLM-based approaches, we test GenTKG~\cite{gentkg} and align with our model settings, we focus on the effects of 8-shot in-context learning for Llama-2-7b~\cite{llama}, Vicuna-7b~\cite{vicuna}, and GPT-NeoX-20B~\cite{neox}. In addition to these, we also include the rule-based method TLogic~\cite{tlogic} in our comparison.

\subsection{Evaluation Protocol}

We acknowledge that, at the metric level, notable distinctions exist between LLM-based methods and embedding-based approaches. The latter proves advantageous as it can furnish a precise ranking of all entities in the graph for a query presented in the form of $(s, q, ?)$, facilitating the calculation of metrics like Mean Reciprocal Rank~\cite{pairre, triplere}. However, for LLM-based methods, we can only furnish the ranking of a predetermined number of candidates, relying on the probabilities of output paths from the open-source model~\cite{icl_tkgc}. This is in contrast to obtaining the ranking of all entities in the graph. This constraint stems from the inability to compel the model to remember all entities directly, and it introduces impractical search costs. Consequently, we choose to report relatively accurate Hits@1, Hits@3, and Hits@10~\cite{rotate}. Furthermore, we align with the perspective outlined in~\cite{ode_tkgc, protocol} that directly excluding all other valid candidates to a specific query in a filtering setting is not entirely reasonable. Additionally, given that the proprietary LLMs we employ for comparison lack the opportunities to output ranking lists, we report raw metrics without loss of generality.\footnote{Supplementary details are in Appendix~\ref{sec: supplementary details}.}

\subsection{Main Results}

As shown in Tbl.~\ref{tab: link prediction results PART1}, Llama-2-7b-CoH and Vicuna-7b-CoH achieves results that surpass or are comparable to the state-of-the-art across multiple metrics under raw setting. Significantly, on the ICEWS05-15 and YAGO datasets, Vicuna-7b-CoH shows an improvement of 3.3\% and 1.9\% in the Hits@1 metric compared to the current best models. We observe that on the YAGO dataset, the 8-shot ICL performance of GPT-NeoX-20B, Llama-2-7b, and vicuna-7b is not significantly worse than Llama-2-7b-CoH. However, there is a noticeable gap on the ICEWS14 series datasets, even falling behind embedding-based models. We also report the metrics under the time-aware filtered setting in Tbl.~\ref{tab: link prediction results PART2}, where Llama-2-7b-CoH outperforms the previous best-performing TiRGN model by 4.1 percentage points in the Hits@1 on YAGO and also exhibits a substantial advantage on ICEWS05-15 and ICEWS18. The relative performance of the model remains generally consistent under both settings.

\section{Analysis}
\subsection{Effective Stucture-based Augmentation}

To assess the efficacy of the structure-augmented history modeling strategy, we conduct comprehensive ablation experiments on all used datasets, employing Hits@1 as the evaluation criterion. For comparison, we exclude entity-augmented and relation-augmented histories during both the fine-tuning and inference phases, relying solely on schema-matching history for predictive determination. The results of the ablation studies are depicted in Tbl.~\ref{tab: ablation on augmentation}, enabling a clear analysis that structure-augmented history is beneficial for both forward and backward inference.

Illustrating with a practical case, when reasoning about the quadruple (\textit{Economist (United Kingdom), Criticize or denounce, ?, 6960}), due to schema-matching history capturing only a historical fact (\textit{Economist (United Kingdom), Criticize or denounce, Silvio Berlusconi, 120}), this leads to an incorrect inference of \textit{Afghanistan}. However, the entity-augmented history contains multiple instances of \textit{Economist (United Kingdom)} linked through the \textit{Make statement} relation to \textit{United Kingdom}. This similar behavior guides the model to output the correct answer \textit{United Kingdom}. Thus, supplementation enhances to some extent the expression of structured information related to the central node, thereby aiding LLM in making more accurate predictions beyond simply relying on the ground truth history.

\subsection{Effect of Introducing Reverse Logic}

\begin{table*}[!t]
    \begin{center}
      \resizebox{1.0\textwidth}{!}{
    \begin{tabular}{l|ccc|ccc|ccc|ccc} 
      \hline % <-- Toprule here
      \multicolumn{1}{l}{Datasets} & \multicolumn{3}{|c}{\textbf{ICEWS14}} &  \multicolumn{3}{|c}{\textbf{ICEWS05-15}} & \multicolumn{3}{|c}{\textbf{ICEWS18}} & \multicolumn{3}{|c}{\textbf{YAGO}}\\
      \hline 
       \  & Forward & Backward & Overall & Forward & Backward & Overall & Forward & Backward & Overall & Forward & Backward & Overall  \\
       \hline 
       Llama-2-7b-CoH w/o aug & 0.353 & 0.297 & 0.325 & 0.400 & 0.357 & 0.379 & 0.226 & 0.196 & 0.211 & 0.555 & 0.491 & 0.523 \\
       Llama-2-7b-CoH & 0.370 & 0.308 & 0.339 & 0.408 & 0.359 & 0.383 & 0.236 & 0.204 & 0.220 & 0.560 & 0.491 & 0.526 \\
       \hline 
       $\Delta$ &  \textcolor{mydarkgreen}{4.8\%} & \textcolor{mydarkgreen}{3.7\%} & \textcolor{mydarkgreen}{4.3\%} & \textcolor{mydarkgreen}{2.0\%} & \textcolor{mydarkgreen}{0.6\%} & \textcolor{mydarkgreen}{1.1\%} & \textcolor{mydarkgreen}{4.4\%} & \textcolor{mydarkgreen}{4.1\%} & \textcolor{mydarkgreen}{4.3\%} & \textcolor{mydarkgreen}{0.9\%} & \textcolor{mydarkgreen}{0.0\%} & \textcolor{mydarkgreen}{0.6\%}
       \\
      \hline 
    \end{tabular}}
    \caption{Ablations on the structure-based history augmentation. We also report \textbf{Hits@1} metric. The strategy has achieved comprehensive improvement in bi-directional forecasting. 
    }
    \label{tab: ablation on augmentation}
    \end{center}
\end{table*}

\begin{table*}[htbp]
    \begin{center}
      \resizebox{1.0\textwidth}{!}{
    \begin{tabular}{l|ccc|ccc|ccc|ccc} 
      \hline % <-- Toprule here
      \multicolumn{1}{l}{Datasets} & \multicolumn{3}{|c}{\textbf{ICEWS14}} &  \multicolumn{3}{|c}{\textbf{ICEWS05-15}} & \multicolumn{3}{|c}{\textbf{ICEWS18}} & \multicolumn{3}{|c}{\textbf{YAGO}}\\
      \hline 
       \  & Forward & Backward & Overall & Forward & Backward & Overall & Forward & Backward & Overall & Forward & Backward & Overall  \\
       \hline 
       Llama-2-7b-CoH w/o rq & 0.367 & 0.298 & 0.333 & 0.396 & 0.343 & 0.369 & 0.238 & 0.188 & 0.213 & 0.560 & 0.489 & 0.524 \\
       Llama-2-7b-CoH & 0.370 & 0.308 & 0.339 & 0.408 & 0.359 & 0.383 & 0.236 & 0.204 & 0.220 & 0.560 & 0.491 & 0.526 \\
       \hline 
       $\Delta$ &  \textcolor{mydarkgreen}{0.8\%} & \textcolor{mydarkgreen}{3.4\%} & \textcolor{mydarkgreen}{1.8\%} & \textcolor{mydarkgreen}{3.0\%} & \textcolor{mydarkgreen}{4.7\%} & \textcolor{mydarkgreen}{3.8\%} & \textcolor{red}{0.8\%} & \textcolor{mydarkgreen}{8.5\%} & \textcolor{mydarkgreen}{3.3\%} & \textcolor{mydarkgreen}{0.0\%} & \textcolor{mydarkgreen}{0.4\%} & \textcolor{mydarkgreen}{0.4\%}
       \\
      \hline 
    \end{tabular}}
    \caption{Ablations on the incorporation of reciprocal quadruples when fine-tuning. We report Hits@1 on four datasets. Rising and falling trends are indicated by \textcolor{mydarkgreen}{green} and \textcolor{red}{red} respectively. In order to more clearly observe differences, we use historical chain with a length of 30 on the ICEWS18 dataset, while for other datasets, this value is set to 10.
    }
    \label{tab: ablation on direction}
    \end{center}
\end{table*}

We conduct a comprehensive ablation experiment for the introduction of reverse quadruples in the fine-tuning phase. Considering the difficulty of ICEWS18 dataset, we set the length of the history chain to 30, and we set this value to 10 on the other datasets. We use the ordinary prompt as a comparison to verify the effect of the reverse data introduction. The results are demonstrated in Tbl.~\ref{tab: ablation on direction}, where Llama-2-7b-CoH~(w/o rq) indicates that no reverse quadruples are added during the fine-tuning phase. We can see that all the results show an upward trend except for a slight dip in the forward inference on the ICEWS18 dataset. Therefore, we can argue that the inclusion of reverse logic in the fine-tuning stage is not only beneficial to alleviate the curse of reversal in structured knowledge reasoning, but also largely harmless to forward reasoning.

We still give a comparison of three proposed prompt styles in Tbl.~\ref{tab: three prompts}. We observe that ordinary and text-aware strategies always lead to better results, so we believe that consistency in preserving the inflectional position of different structured quadruples during fine-tuning is more critical.

\begin{table}[t]
    \centering
    \resizebox{1.0\columnwidth}{!}{
        \begin{tabular}{l|cccc}
            \hline
            Strategy & YAGO & ICEWS14 & ICEWS05-15 & ICEWS18 \\
            \hline
            Ordinary & \textbf{0.526} & \textbf{0.339} & \textbf{0.383} & 0.209 \\
            Text-aware & 0.525 & 0.333 & 0.382 & \textbf{0.214} \\
            Position-aware & 0.525 & 0.330 & 0.381 & 0.213 \\ \hline
            
        \end{tabular}
    }
    \vspace{-0.1cm}
    \caption{Overall Hits@1 metrics for three utilized prompt strategies under the raw setting.}
    \label{tab: three prompts}
    \vspace{-0.3cm}
\end{table}

\subsection{Exploration on History Length}
The length of the historical chain $L$ significantly influences prediction outcomes, reflecting the amount of information provided to the LLMs. We conduct experiments with varying history lengths ($L=10, 20, 30, 50$), while maintaining other settings constant. We choose the \textit{ordinary prompt} for incorporating reverse quadruples and harness entity-augmented and relation-augmented quadruples to enrich historical facts.

\begin{figure}
    \centering
    \vspace{-0.2cm}
    \includegraphics[width=\linewidth]{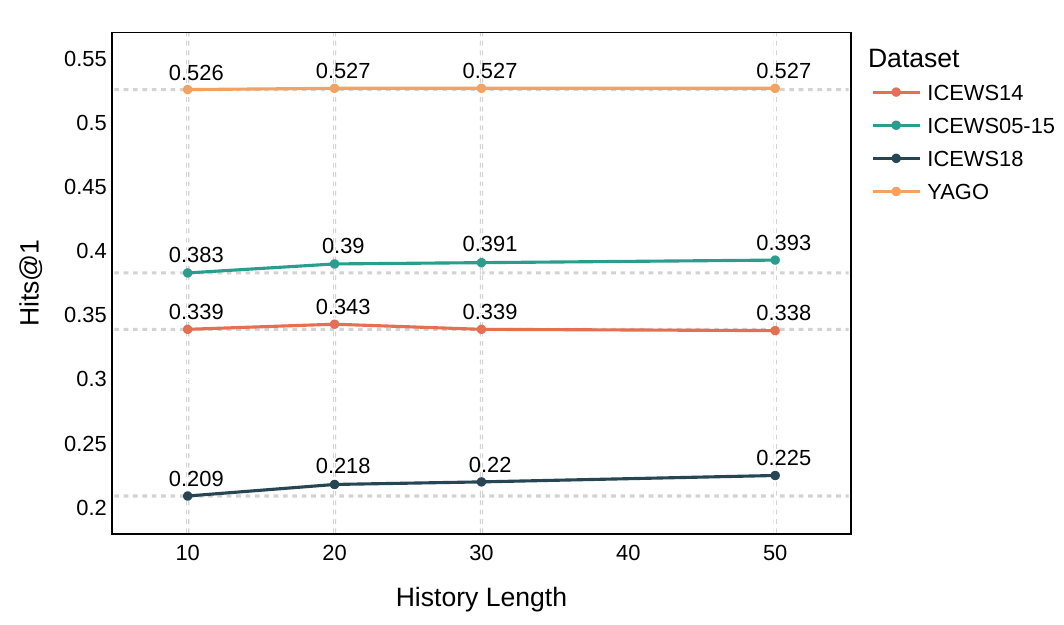}
    \caption{The evolution pattern of the Hits@1 metric across four utilized datasets concerning the history length $L$.}
    \label{fig: history length}
\end{figure}

As illustrated in Fig.~\ref{fig: history length}, except the ICEWS14 dataset, on other datasets, the Hits@1 metric exhibits an upward trend followed by stabilization as $L$ increases. We calculate the average length of schema-matching history for each query in the test sets of four datasets. For the ICEWS14 dataset, this value is 30.05, significantly lower than the other datasets. On the ICEWS05-15 dataset, this value is 56.95. Consequently, an excessively long required history length may negatively impact the reasoning of LLM due to interference from numerous historical quadruples used for padding. However, even with a smaller input cost (i.e., smaller $L$) on the ICEWS14 dataset, significant effectiveness is already achievable.

\subsection{How Model Size Affects Results}
\begin{table}[t]
    \centering
    \resizebox{1.0\columnwidth}{!}{
        \begin{tabular}{l|cccc}
            \hline
            Model & YAGO & ICEWS14 & ICEWS05-15 & ICEWS18 \\
            \hline
            Llama-2-7b-CoH & 0.527 & 0.343 & 0.390 & 0.218 \\
            Llama-2-13b-CoH & 0.526 & 0.343 & 0.392 & 0.210 \\ 
            Vicuna-33b & 0.530 & 0.338 & 0.390 & 0.216 \\
            \hline
            
        \end{tabular}
    }
    \vspace{-0.1cm}
    \caption{Overall Hits@1 metrics on different model sizes.}
    \label{tab: model size}
    \vspace{-0.5cm}
\end{table}

\begin{table*}[htbp]
    \begin{center}
      \resizebox{1.0\textwidth}{!}{
    \begin{tabular}{l|ccc|ccc|ccc|ccc} 
      \hline % <-- Toprule here
      \multicolumn{1}{l}{Datasets} & \multicolumn{3}{|c}{\textbf{ICEWS14}} &  \multicolumn{3}{|c}{\textbf{ICEWS05-15}} & \multicolumn{3}{|c}{
      
      \textbf{ICEWS18}} & \multicolumn{3}{|c}{\textbf{YAGO}}\\
      \hline 
       \  & Forward & Backward & Overall & Forward & Backward & Overall & Forward & Backward & Overall & Forward & Backward & Overall  \\
       \hline 
       GPT-3.5-turbo & 0.260 & 0.158 & 0.209 & 0.157 & 0.177 & 0.167 & 0.079 & 0.070 & 0.075 & 0.496 & 0.441 & 0.481 \\
       GPT-4~\cite{gpt4report} & \textbf{0.298} & \textbf{0.233} & \textbf{0.266} & 0.293 & 0.260 & 0.277 & 0.096 & 0.092 & 0.094 & \textbf{0.510} & \textbf{0.484} & \textbf{0.497} \\
       Qwen-72B-Chat~\cite{QWENreport} & 0.279 & 0.216 & 0.248 & \textbf{0.357} & \textbf{0.343} & \textbf{0.350} & \textbf{0.159} & \textbf{0.148} & \textbf{0.154} & 0.499 & 0.463 & 0.481 \\
      \hline 
    \end{tabular}}
    \caption{The performance of some powerful commercial models on 1000 randomly selected test samples in each dataset.}
    \label{tab: commercial_llms}
    \end{center}
\end{table*}
In this section, we explore how model size of LLMs affects performance in TKGC. We choose Llama-2-13b and Vicuna-33b as comparison and consider leveraging total history length with $L=20$, and both add inverse quadruples and structure-based augmentation data for fine-tuning. The results, as shown in Tbl.~\ref{tab: model size}, depict that these three sizes models achieve very similar results in Hits@1. Unusually, Hits@1 on ICEWS18 dataset decreases by 3.7\% and 0.9\% compared to Llama-2-7b-CoH. We point out that increasing the size of the model is a relatively inefficient approach in the context of temporal logical reasoning. Larger models do not necessarily result in a better understanding of interactive information along the temporal chain. This leads us to explore data-centric approaches and improvements in the inherent reasoning limitations of LLMs, such as catastrophic forgetting and the curse of reversibility.

\subsection{Performance of Commercial LLMs}

In this section, we test the effectiveness of three powerful commercial LLMs on the TKGC task, aiming to explore the performance differences after multi-task instruction fine-tuning and Reinforcement Learning from Human Feedback~(RLHF). We provide the same 8-shot ICL prompt samples for each of the three models on different datasets, as detailed in the appendix. For the test data, we randomly select 1000 queries for both directions on each dataset. Since these models do not provide output probabilities, we only present the most accurate exact match metric, equivalent to the Hits@1 metric under the raw setting. After confirming that there are no fine-tuning on TKGC task and related datasets in the available technical reports~\cite{gpt4report,QWENreport}, we consider this comparison to be relatively fair. 

The evaluation results are shown in Tbl.~\ref{tab: commercial_llms}. Firstly, we can observe that Qwen-72B-Chat is able to achieve performance comparable to or surpass GPT-4. In contrast, the performance of GPT-3.5-turbo is not satisfactory. We are currently observing that the few-shot capabilities of Qwen-72B-Chat on the MMLU evaluation set are approaching those of GPT-4 and surpassing the performance of GPT-3.5-turbo. This eliminates a significant bias in terms of language tendency. On the other hand, we demonstrate that chat models, carefully fine-tuned and applying RLHF, exhibit superior performance in TKGC tasks. However, when we compare the results of Tbl.~\ref{tab: commercial_llms} and Tbl.~\ref{tab: link prediction results PART1}, we can observe that the 8-shot ICL capability of commercial LLMs is still significantly lower on the ICEWS series dataset compared to the capabilities of Llama-2-7b-CoH, while the difference is not substantial on the YAGO. This is because YAGO is a dataset biased towards common knowledge, and therefore, commercial LLMs may already be familiar with a considerable number of rules. However, the reasoning in the ICEWS series news dataset emphasizes the interaction and evolutionary information of nodes in the graph rather than relying on textual features. This results in commercial LLMs underperforming in ICL, as they struggle to effectively capture the evolutionary patterns along historical chains and utilize augmented structure-based knowledge. 

\section{Conclusion}

In this study, we conceptualize Temporal Knowledge Graph Completion (TKGC) as a dual-process of fine-tuning and generative procedures of LLMs along the historical chain. Our comprehensive exploration extends to the perceptual capabilities of LLMs to interpret graph modality and structured knowledge. To augment the understanding of central nodes by LLMs, we devise a series of structure-based enhanced quadruples, premised on entity nodes and relations. Furthermore, we address the reversal curse in LLMs by introducing reverse logic data. Our approach surpasses or equals the performance of existing models. We also offer in-depth analysis of the factors influencing the model's inference capabilities, highlighting the contributions of the proposed fine-tuning pipeline. Our findings still indicate that models tend to fit better with extended historical data. However, the model's size is a less significant factor, and the subpar performance of commercial LLMs suggests that RLHF in broad domains may not necessarily enhance inference tasks. We posit that our discoveries will stimulate the reciprocal advancement of LLMs and TKGC.

% \section{Ethics Consideration}
% I ensure that every author reads and consciously adheres to the ACL Code of Ethics. The ethical concerns in our experiments mainly stem from the use of open-source LLMs and the employment of news and common knowledge datasets involving political people and events. Consequently, the process of generating answers may uncontrollably produce incorrect results that may be misunderstood to a certain degree by specific groups of people. However, our results are only used for recording the correctness of inference and will not be disclosed or disseminated.

% \section{Limitaions}

% Our research still has many limitations. The integration of TKGs and LLMs has some inherent flaws. For example, whether LLMs have previously stored the knowledge in these widely-used datasets in the form of unstructured text, and a considerable portion of the queries in the benchmark of TKGs cannot be answered correctly by known events. These factors limit the accuracy and scalability of the study. Moreover, in exploring the impact of model size on experimental results, we have not yet explored models larger than 33b parameters. Although current data suggests that model size does not bring about positive gains in inference, there is still the possibility of qualitative changes due to quantitative changes. Our model selection is also limited to the \textit{Llama} and \textit{Vicuna} series, without extending to other open-source models.

\bibliography{anthology,custom}
\bibliographystyle{acl_natbib}

\newpage
\appendix
\label{sec:appendix}

\begin{table*}[!t]
 \begin{center}
 \resizebox{0.9\textwidth}{!}{

    \begin{tabularx}{1.0\linewidth}{lX}
    \hline 
    \textbf{Section} & \textbf{Prompt} \\
    \hline
    Instruction & Given contexts consisting of multiple quadruplets in the form of \{time\}: [\{subject\}, \{relation\}, \{object\}], please predict the missing entity in the query quadruplet \{time\}: [\{subject\}, \{relation\}, ] in the end. \\
    \hline
    Input & 295: [[Victor\_Ponta, Make\_statement, Romania] \\
&296: [Victor\_Ponta, Make\_statement, North\_Atlantic\_Treaty\_Organization]\\ 
&296: [Victor\_Ponta, Make\_statement, Romania]\\
&300: [Victor\_Ponta, Make\_statement, Viorel\_Hrebenciuc] \\
&301: [Victor\_Ponta, Make\_statement, Romania] \\
&302: [Victor\_Ponta, Make\_statement, Romania] \\
&303: [Victor\_Ponta, Make\_statement, National\_Liberal\_Party\_(Romania)] \\
&303: [Victor\_Ponta, Make\_statement, Romania] \\
&304: [Victor\_Ponta, Make\_statement, Romania] \\
&307: [Victor\_Ponta, Make\_statement, Representatives\_(Romania)] \\
&Query:\\
&308: [Victor\_Ponta, Make\_statement, ] \\
    \hline
    Output &  The missing entity of query quadruplet is Romania. \\
    \hline
    \end{tabularx}
}
\caption{Prompt design using text only.}
\label{tab:IT}
\end{center}
\end{table*}

\begin{table*}[!t]
 \begin{center}
 \resizebox{0.9\textwidth}{!}{

    \begin{tabularx}{1.0\linewidth}{lX}
    \hline 
    \textbf{Section} & \textbf{Prompt} \\
    \hline
    Instruction & Given contexts consisting of multiple quadruplets in the form of \{time\}: [\{subject\}, \{relation\}, \{label\}.\{object\}], please predict the missing entity in the query quadruplet \{time\}: [\{subject\}, \{relation\}, ] in the end. \\
    \hline
    Input & 295: [[Victor\_Ponta, Make\_statement, 0.Romania] \\
&296: [Victor\_Ponta, Make\_statement, 1.North\_Atlantic\_Treaty\_Organization]\\ 
&296: [Victor\_Ponta, Make\_statement, 0.Romania]\\
&300: [Victor\_Ponta, Make\_statement, 2.Viorel\_Hrebenciuc] \\
&301: [Victor\_Ponta, Make\_statement, 0.Romania] \\
&302: [Victor\_Ponta, Make\_statement, 0.Romania] \\
&303: [Victor\_Ponta, Make\_statement, 3.National\_Liberal\_Party\_(Romania)] \\
&303: [Victor\_Ponta, Make\_statement, 0.Romania] \\
&304: [Victor\_Ponta, Make\_statement, 0.Romania] \\
&307: [Victor\_Ponta, Make\_statement, 4.Representatives\_(Romania)] \\
&Query:\\
&308: [Victor\_Ponta, Make\_statement, ] \\
    \hline
    Output &  The missing entity of query quadruplet is 0.Romania. \\
    \hline
    \end{tabularx}
}
\caption{Prompt design using text and id.}
\label{tab:IT with id}
\end{center}
\end{table*}

\section{Appendix}
\subsection{Instruction Used by CoH}
\label{sec:app coh}

In this section, we provide a comprehensive design for the prompt, including versions that utilize only entity text~(Tbl.~\ref{tab:IT}) and versions identified by number id~(Tbl.~\ref{tab:IT with id}).

\subsection{Prompt for 8-shot ICL}
\label{sec:app icl}

We design different prompts on different datasets to test the ability of different models to perform ICL on the TKGC task. We show the prompt template on the ICEWS18 dataset as a concrete example, as shown in the Tbl.~\ref{tab:prompt for icl}.

\begin{table*}[!t]
 \begin{center}
 \resizebox{0.95\textwidth}{!}{

    \begin{tabularx}{1.0\linewidth}{X}
    \hline 
    \multicolumn{1}{c}{\textbf{8-shot Prompt}} \\
    \hline
    You must be able to correctly predict the next \{object\} from a given text consisting of multiple quadruplets in the form of "\{time\}:[\{subject\}, \{relation\}, \{object\}]" and the query in the form of "\{time\}:[\{subject\}, \{relation\}," in the end.\\
Example 1: 3864: [Police (Malaysia), Confiscate property, Malaysia] 4272: [Police (Malaysia), Confiscate property, Malaysia] 4944: [Police (Malaysia), Confiscate property, Malaysia] 5952: [Police (Malaysia), Confiscate property, Malaysia] 6072: [Police (Malaysia), Confiscate property, Malaysia] 6192: [Police (Malaysia), Confiscate property, Indonesia] 6288: [Police (Malaysia), Confiscate property, Citizen (Malaysia)] 6336: [Police (Malaysia), Confiscate property, Citizen (Malaysia)] \\
Example 2: 6408: [Police (India), Accuse, Criminal (India)] 6408: [Police (India), Accuse, Student (India)] 6408: [Police (India), Accuse, Citizen (India)] 6432: [Police (India), Accuse, Criminal (India)] 6456: [Police (India), Accuse, Inspector General (India)] 6456: [Police (India), Accuse, Citizen (India)] 6456: [Police (India), Accuse, Children (India)] 6456: [Police (India), Accuse, Women (India)] \\
Example 3: 6120: [China, Reject, India] 6336: [China, Reject, United States] 6384: [China, Reject, United States] 6432: [China, Reject, Naval (United States)] 6432: [China, Reject, Donald Trump] 6432: [China, Reject, United States] 6456: [China, Reject, Donald Trump] 6456: [China, Reject, United States]\\
Example 4: 6408: [Shinzo Abe, Consult, North Korea] 6408: [Shinzo Abe, Consult, Head of Government (South Korea)] 6432: [Shinzo Abe, Consult, Kim Jong-Un] 6432: [Shinzo Abe, Consult, Moon Jae-in] 6432: [Shinzo Abe, Consult, Hassan Rouhani] 6432: [Shinzo Abe, Consult, Donald Trump] 6432: [Shinzo Abe, Consult, UN General Assembly] 6456: [Shinzo Abe, Consult, Donald Trump] \\
Example 5: 5568: [Joao Lourenco, Make a visit, Germany] 5592: [Joao Lourenco, Make a visit, Germany] 5616: [Joao Lourenco, Make a visit, Germany] 5736: [Joao Lourenco, Make a visit, Angola] 5976: [Joao Lourenco, Make a visit, China] 6408: [Joao Lourenco, Make a visit, United States] 6720: [Joao Lourenco, Make a visit, China] 6768: [Joao Lourenco, Make a visit, China] \\
Example 6: 5208: [Saudi Arabia, Demand, Foreign Affairs (Canada)] 5256: [Saudi Arabia, Demand, Student (Saudi Arabia)] 5256: [Saudi Arabia, Demand, Canada] 5304: [Saudi Arabia, Demand, Student (Saudi Arabia)] 5760: [Saudi Arabia, Demand, Sudan] 6288: [Saudi Arabia, Demand, Citizen (Saudi Arabia)] 6792: [Saudi Arabia, Demand, Jamal Khashoggi] 6816: [Saudi Arabia, Demand, Jamal Khashoggi] \\
Example 7: 4248: [Wei Fenghe, Express intent to cooperate, James Mattis] 6552: [Wei Fenghe, Consult, Department of Defense] 6552: [Wei Fenghe, Halt negotiations, James Mattis] 6960: [Wei Fenghe, Consult, James Mattis] 6960: [Wei Fenghe, Meet at a 'third' location, James Mattis] 6960: [Wei Fenghe, Make a visit, ASEAN Defense Ministers] 6960: [Wei Fenghe, Engage in negotiation, James Mattis] 6960: [Wei Fenghe, Halt negotiations, James Mattis] \\
Example 8: 6936: [Police (India), Arrest, detain, or charge with legal action, Student (India)] 6960: [Police (India), Arrest, detain, or charge with legal action, Men (India)] 6960: [Police (India), Arrest, detain, or charge with legal action, Criminal (India)] 6960: [Police (India), Arrest, detain, or charge with legal action, Children (India)] 6960: [Police (India), Arrest, detain, or charge with legal action, Citizen (India)] 6960: [Police (India), Arrest, detain, or charge with legal action, Student (India)] 6960: [Police (India), Arrest, detain, or charge with legal action, Parkash Singh Badal] 6960: [Police (India), Arrest, detain, or charge with legal action, Women (India)] \\
    \hline
    \end{tabularx}
}
\caption{8-shot ICL prompt design on ICEWS18.}
\label{tab:prompt for icl}
\end{center}
\end{table*}

\subsection{Supplementary Details}\label{sec: supplementary details}
In this section, we describe the supplementary settings of our experiments. The open-source models mainly used are \textit{llama-2-7b}, \textit{llama-2-13b}, \textit{vicuna-7b-v1.5}, and \textit{vicuna-33b-v1.3}. The key search parameters during fine-tuning and inference are shown in the Tbl.~\ref{tab: parameters}. Our main experiments in Tbl.~\ref{tab: link prediction results PART1} and Tbl.~\ref{tab: link prediction results PART2} run on 4*NVIDIA GeForce RTX 4090, and studies of \textit{vicuna-33b-v1.3} run on 4*NVIDIA A100-SXM-80G.

\begin{table*}[!t]
 \begin{center}
 \resizebox{0.95\textwidth}{!}{

    \begin{tabularx}{1.0\linewidth}{lX}
    \hline 
    \textbf{Parameter} & \textbf{Candidates} \\
    \hline
    batch\_size & 4, 8 \\
    lora\_rank & 8, 32 \\
    lora\_dropout & 0.1 \\ 
    lora\_target\_modules & \{q\_proj,k\_proj,v\_proj,o\_proj\},\ \{q\_proj, k\_proj\} \\
    lora\_alpha & 16\\
    truncation\_length & 3000 \\
    
    $L$ & 10, 20, 30, 40, 50\\
    single\_step\_inference\_candidate & 10 \\
    \hline
    \end{tabularx}
}
\caption{Parameter search space.}
\label{tab: parameters}
\end{center}
\end{table*}

\end{document}